\documentclass{article}

\usepackage{microtype}
\usepackage{graphicx}
\usepackage{subfigure}
\usepackage{booktabs} 

\usepackage{hyperref}

\usepackage[accepted]{icml2023}

\usepackage{amsmath}
\usepackage{amssymb}
\usepackage{mathtools}
\usepackage{amsthm}

\usepackage[capitalize,noabbrev]{cleveref}

\theoremstyle{plain}
\newtheorem{theorem}{Theorem}[section]

\newtheorem{corollary}[theorem]{Corollary}
\theoremstyle{definition}
\newtheorem{definition}[theorem]{Definition}

\theoremstyle{remark}
\newtheorem{remark}[theorem]{Remark}

\usepackage[textsize=tiny]{todonotes}

\usepackage{amsfonts}
\usepackage{amsmath}
\usepackage{booktabs}
\usepackage{amsthm}

\newcommand\numberthis{\addtocounter{equation}{1}\tag{\theequation}}
\usepackage{bbm}
\usepackage{bm}
\usepackage{xcolor}
\usepackage{amsthm}

\icmltitlerunning{Learning Prescriptive ReLU Networks}

\begin{document}

\twocolumn[
\icmltitle{Learning Prescriptive ReLU Networks}



\icmlsetsymbol{equal}{*}

\begin{icmlauthorlist}
\icmlauthor{Wei Sun}{ibm}
\icmlauthor{Asterios Tsiourvas}{mit}

\end{icmlauthorlist}

\icmlaffiliation{mit}{Operations Research Center, Massachusetts Institute of Technology, Cambridge, MA, USA}
\icmlaffiliation{ibm}{IBM Research, Yorktown Heights, NY, USA}

\icmlcorrespondingauthor{Asterios Tsiourvas}{atsiour@mit.edu}

\icmlkeywords{ReLU Networks, Causal ML, Prescriptive Trees, Constraints, Explainable AI}

\vskip 0.3in
]



\printAffiliationsAndNotice{}  

\begin{abstract}
We study the problem of learning optimal policy from a set of discrete treatment options using observational data. We propose a piecewise linear neural network model that can balance strong prescriptive performance and interpretability, which we refer to as the \emph{prescriptive ReLU network}, or \emph{P-ReLU}. We show analytically that this model (i) partitions the input space into disjoint polyhedra, where all instances that belong to the same partition receive the same treatment, and (ii) can be converted into an equivalent prescriptive tree with hyperplane splits for interpretability. We demonstrate the flexibility of the P-ReLU network as constraints can be easily incorporated with minor modifications to the architecture. Through experiments, we validate the superior prescriptive accuracy of P-ReLU against competing benchmarks. Lastly, we present examples of interpretable prescriptive trees extracted from trained P-ReLUs using a real-world dataset,  for both the unconstrained and constrained scenarios. 
\end{abstract}

\section{Introduction}

The problem of determining the best  treatment option based on an individual's characteristics from data  is a central problem across many domains, e.g., ad targeting  \cite{li2010contextual},  precision medicine \cite{ekins2019exploiting}, and contextual pricing \cite{biggs2021loss}. In many such settings, only {observational data} is available. That is, for each observation, we have the features that describe the instance (e.g., a patient or a customer), the treatment prescribed, and the outcome of the chosen action. As its name suggests, the data are collected passively and we do not control the historic administration of treatments. In particular, the {counterfactuals}, i.e., outcomes associated with alternative treatments, are not known. Besides the issue of missing counterfactuals, another challenge of policy learning stems from the need for interpretability - which is critical for adoption in practice.  Complex and opaque policies not only make implementation cumbersome but also are  difficult for humans to understand and trust.  

In recent years, there has been a stream of nascent 
 research focusing on learning interpretable optimal policy from data in the form of 
\emph{prescriptive trees} \cite{biggs2021model,amram2022optimal,zhou2022offline}, since the tree structure is visually easy to understand \cite{OMT}. In contrast to standard decision trees for prediction tasks, prescriptive trees are constructed to provide policies that optimize a given objective (e.g., pricing strategies that maximize revenue, medical treatment to minimize a certain symptom). More specifically,  
all samples in a leaf node of a prescriptive tree are prescribed a same treatment, and each path from the root node to a leaf node  corresponds to a policy. In particular, \citet{kallus2017recursive} and \citet{bertsimas2019optimal} propose an  integrated approach whereby a counterfactual prediction model is embedded within the tree-building algorithm, combining the tasks of estimating the counterfactuals and learning the optimal prescription policy in a single step. However, to maintain tractability, 
 the counterfactual model which needs to be evaluated at every node of a tree is restricted to constant or linear functions. This limitation potentially leads to significant model misspecification when the underlying outcome function takes a more complex form.  

The goal of this paper is to introduce a model which is expressive enough to achieve superior prescriptive performance and can also be flexible to provide interpretable policy. We propose a  neural network model with ReLU activation functions, which we refer to as the \emph{prescrptive ReLU} network or \emph{P-ReLU} in short. Our key contributions are: 

\begin{itemize}

\item We propose an integrated approach that estimates the counterfactuals and learns the optimal policies simultaneously. We utilize a ReLU neural network that can approximate any continuous function to model the counterfactuals, instead of the constant or linear functions used in \citet{kallus2017recursive} and \citet{bertsimas2019optimal}. Moreover, existing tree-based prescriptive models to date partition the input space using axis-aligned splits. P-ReLU networks partition the input space into general polyhedra, contributing to its superior model performance in terms of prescriptive accuracy.

    \item We show analytically (i) that P-ReLU partitions the input space into disjoint convex polyhedra, where all instances that belong to the same partition receive the same treatment, and (ii) that every P-ReLU network can be transformed into an equivalent  prescriptive tree with hyperplane splits. 
    \item To generate meaningful policy, it is often critical to incorporate constraints (e.g., if a patient exhibits certain symptoms, some treatments are prohibited). 
    We show that a P-ReLU network can be easily modified to incorporate such constraints. In addition, constrained P-ReLUs can be trained efficiently using gradient descent, highlighting the flexibility of the method. 
    \item We experiment with the proposed method on both synthetic and real-world datasets. 
    P-ReLUs exhibit superior  prescriptive accuracy over competing benchmarks. To demonstrate its potential at generating interpretable policy, we extract corresponding prescriptive trees from trained networks on both unconstrained and constrained scenarios. 
\end{itemize}

\section{Related Literature}

The problem of learning optimal policy is closely related to estimating heterogeneous treatment effect in the causal inference literature. One approach is called meta-learners \cite{kunzel2019metalearners}, which uses base-learners to estimate the conditional expectations of the outcomes  for control and treatment groups, and then takes the difference between these estimates. S-Learner fits a model to estimate outcomes by using all features, without giving the treatment indicator a special role  tends to produce biased estimators. T-learner, also known as regress-and-compare (R\&C) method \cite{jo2021learning}, splits the training data by treatments, fits a regression model in each sub-population. By building models on separate datasets which  decreases the efficiency of learning, it can also be difficult to learn behaviors that are common across treatment groups. Other meta-learners include doubly-robust estimator \cite{dudik2011doubly}, X-Learner \cite{kunzel2019metalearners} and R-Learner \cite{nie2021quasi}, which require propensity scores  as an additional input.
There are also tree-based methods in the causal literature to learn the treatment effect:  \citealp{athey2016recursive} use a recursive splitting procedure on the feature space to construct causal trees (CTs), where the average treatment effect (i.e., the difference between the sample mean in the treated and  control group) is used as the prediction for that leaf. 
Later \citealp{wager2018estimation}  extend it to causal forests (CFs).  Building on the ideas of causal trees, \citealp{wager2018estimation} adapt random forests, boosting, and multivariate adaptive regression splines for treatment effect estimation. As this body of work  focuses on quantifying treatment effects, it does not explicitly consider policy optimization which is the key focus of this paper. Moreover,  many of the aforementioned methods (e.g., tree-based methods) are designed only for binary treatments, instead of the multi-treatment setting considered here.

There has been a fast-growing body of research focusing on learning prescriptive trees due to their interpretability. Various tree-building algorithms have been proposed. For instance, the prescriptive tree can be constructed either greedily (\citealp{kallus2017recursive,biggs2021model,zhou2022offline}) or optimally via a mixed integer programming (MIP) formulation (\citealp{bertsimas2019optimal,jo2021learning,amram2022optimal,subramanian2022constrained}). While the latter guarantees global optimality, scalability remains a challenge and such an approach is typically limited to  shallow trees and small datasets with no more than a few thousand samples. Meanwhile, within the tree-constructing procedure, the  prediction step may be explicitly decoupled from  the policy optimization \cite{biggs2021model,amram2022optimal,subramanian2022constrained,zhou2022offline}, or it can also be embedded in the policy generation, yielding an elegant integrated approach  that we are focusing on. Instead of restricting to constant \cite{kallus2017recursive} or  linear \cite{bertsimas2019optimal} functions to model the counterfactuals, 
we utilize a ReLU network that can approximate any continuous function and thus, is more flexible at representing the predicted outcomes.

ReLU networks have attracted significant attention in recent years due to their inherent piecewise linear nature which promotes analytical tractability \cite{montufar2014number,lee2019towards}.  Researchers have utilized these properties for a variety of applications, such as robustness verification \cite{tjeng2017evaluating} and network compression \cite{serra2020lossless}. Additionally, many works have focused on optimizing already trained ReLU networks for downstream tasks, utilizing both mixed-integer optimization techniques \cite{fischetti2018deep,anderson2020strong,de2021scaling} and approximate methods \cite{katz2017reluplex,xu2020reluplex,selfcitation}. Recent studies have also studied their expressive power \cite{arora2016understanding,yarotsky2017error,zou2020gradient} as well as their connection with other machine learning models \cite{lee2019oblique}. While ReLU networks have been typically trained as prediction models,  we expand the stream of literature to the prescriptive setting by proposing a novel ReLU network architecture that  learns optimal policy from observational data.

\section{Methodology}

\subsection{Problem Formulation}

We assume we have $n$ observational data samples $\{(x_t,p_t,y_t)\}_{t=1}^n$, where $x_t \in \mathcal{X} \subset \mathbb{R}^d$ are features that describe instance $t$, $p_t \in [K]:=\{0,1,\dots,K-1\} $ is the policy or treatment taken and $y_t \in \mathbb{R}$ is the observed outcome. We use the convention that small outcomes are preferable.

As a common practice in the causal inference literature, we make the ignorability assumption \cite{hirano2004propensity}. That is,  there were no unmeasured confounding variables that affected both the treatment and the outcome. We seek a function $\pi: \mathcal{X} \to [K] $ that picks the best treatment out of the $K$ possible treatment options for a given set of features $x$. In particular, we want $\pi(x)$ to be both ``optimal'' and ``accurate''. To be precise, \emph{optimality} refers to minimizing \emph{prescription outcome}, i.e., $\mathbb{E}[y(\pi(x))]$, where the expectation is taken over the distribution of outcomes for a given treatment policy $\pi(x)$. We can approximate the expectation in terms of the samples as follows, 
\begin{align}\label{def_prescription}
\sum_{t=1}^n \big(y_t \mathbbm{1}[\pi(x_t) = p_t] + \sum_{ p\neq p_t} \hat{y}_t(p)  \mathbbm{1}[\pi(x_t) = p] \big), 
\end{align}
where $\hat{y}_t(p)$ denotes the estimated outcome with treatment $p$ for instance $t$. 

Meanwhile, \emph{accuracy} refers to minimizing  \emph{prediction error} on the observed data, i.e.,
\begin{align}\label{def_prediction}
\sum_{t=1}^n (y_t - \hat{y}_t(p_t))^2. 
\end{align}

To combine these two objectives, we consider a convex combination of (\ref{def_prescription}) and (\ref{def_prediction}), 
\begin{align}\label{def_loss}
\mu\cdot\mathrm{Prescription\,outcome} + (1-\mu)\cdot \mathrm{Prediction\,error},\end{align}
where $\mu \in [0,1]$ is a hyper-parameter that balances the trade-off between optimality and accuracy. It has been shown in \citealp{bertsimas2019optimal} that this objective outperforms optimizing prescription outcome alone (i.e., $\mu=1$) as  in 
\citealp{kallus2017recursive}. 

The key distinction between  the existing integrated prescriptive tree approach  \cite{kallus2017recursive,bertsimas2019optimal} and our work  is that the former restricts $\pi(x)$ to a tree structure a priori and  the class of models that can be embedded for modeling counterfactuals is limited to constant or linear functions. Our goal is to find a powerful, yet flexible model that can balance strong prescriptive performance and interpretability.  We achieve this by considering the prescription policy to the class of  neural networks with piecewise linear activation functions as basic building blocks.

\begin{figure*}[ht]
\centering
\includegraphics[width=.25\linewidth]{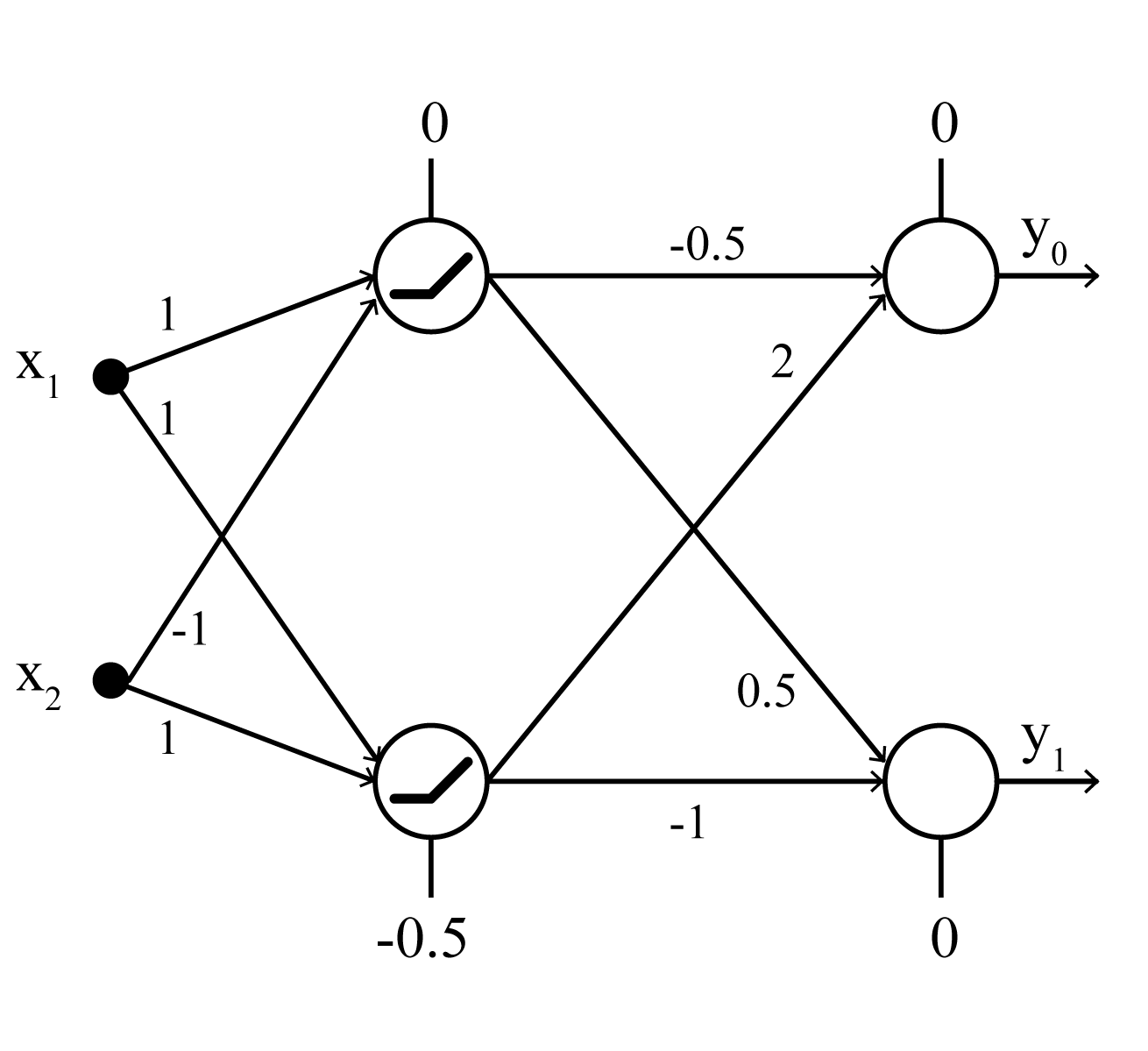}\hfill
\includegraphics[width=.25\linewidth]{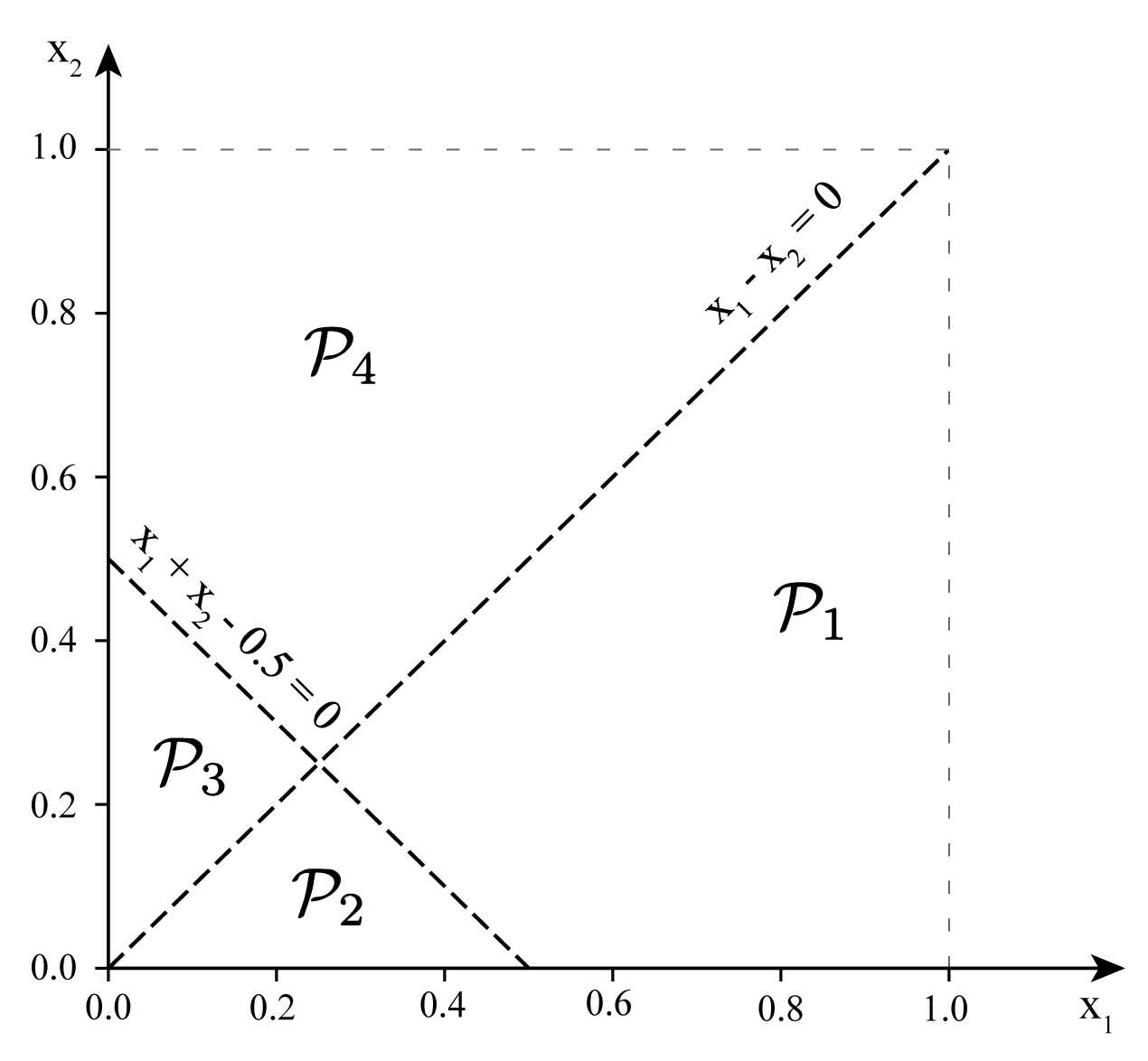}\hfill
\includegraphics[width=.25\linewidth]{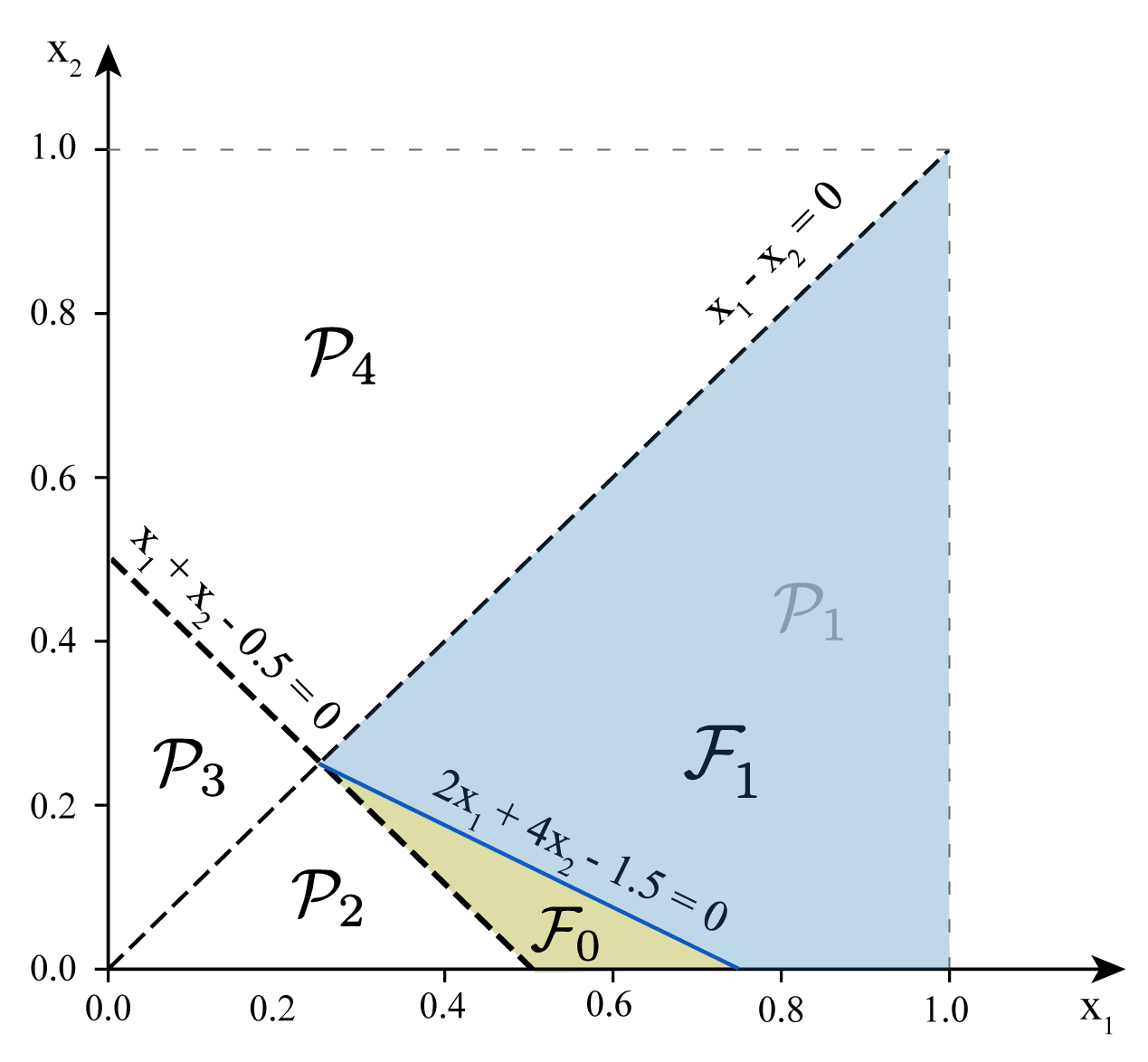}
\caption{A toy example that demonstrates the partition of the input space $\mathcal{X}$ created by the proposed method. (Left) A one-layer, 2 neuron, ReLU neural network. Over the edges, we observe the weights and the biases. (Middle) The partition of 
$\mathcal{X}$ into $4$ disjoint convex polyhedra by the hidden layer of the network. (Right) The final partition of $\mathcal{P}_1$ by the output layer into the treatment regions $\mathcal{F}_0,\mathcal{F}_1$. }
\label{fig:toy_example}
\end{figure*}

\subsection{Network Architecture}

The proposed approach is built on feed-forward neural networks with ReLU activations. We consider the densely connected architecture \cite{huang2017densely}, where each neuron takes the outputs of the neurons in the previous layer as inputs. In the output layer, we restrict the number of neurons to be equal to the number of the total treatments $K$. Each of the $K$ output neurons approximates the outcome under the corresponding treatment. We assign  the treatment with the lowest predicted outcome to each instance.  

Formally, we denote the network as $f_{\theta}: \mathcal{X} \to \mathbb{R}^K$ where $\theta$ is the set of weights. We denote the number of hidden layers as $L$ and the number of neurons at layer $i$ as $N_i$. We also denote the values of the neurons before activation as $x^i \in \mathbb{R}^{N_i}$ and the values of the neurons after activation as $z^i \in \mathbb{R}^{N_i}$. For notational convenience we define $z^0:=x$, $N_0:=d$. The neurons are defined by the weight matrix $W^i \in \mathbb{R}^{N_i \times N_{i-1}} $ and the bias vector $b^i \in \mathbb{R}^{N_i} $. Based on the aforementioned notation, $x^i = W^iz^{i-1}+b^i, z^i = \sigma(x^i)$, where $\sigma(\cdot)$ is the ReLU activation function \cite{nair2010rectified}, i.e., $\sigma(x)=\max \{x,0\}$. 
We also define the activation indicator function $o^{i}_j(x) = \mathbbm{1}[z^i_j > 0]$ for neuron $j$ of layer $i$, where $\mathbbm{1}[\cdot]$ is the indicator function, as well as the activation $o^i(x)=[o^i_1(x),\dots,o^i_{N_i}(x)]^T$ of the layer $i$. Finally, we define the activation pattern $o(x)=[o^1(x),\dots,o^L(x)]^T\in\mathbb{R}^{N}$, where $N := \sum_{i=1}^L N_i$, as the collection of the activation indicator functions for each neuron of the network when the input is $x$. 

The objective of the loss function defined in (\ref{def_loss}) can be rewritten as follows, 
\begin{align*}
    &\mu \cdot \sum_{t=1}^n \big(y_t \mathbbm{1}[\pi_{f_{\theta}}(x_t) = p_t] + \sum_{ p\neq p_t} f_{\theta}(x_t)_p  \mathbbm{1}[\pi_{f_{\theta}}(x_t) = p] \big)\\
    &+(1-\mu) \cdot  \sum_{t=1}^n (y_t -  f_{\theta}(x_t)_{p_t})^2, \label{eq:obj} \numberthis
\end{align*}
\noindent where $f_{\theta}(x_t)_p$ is the predicted outcome to treatment $p$ for instance $x_t$ by the network $f_{\theta}(\cdot)$,  $\pi_{f_{\theta}}(x_t)= \arg \min \limits_{p \in [K]} f_{\theta}(x_t)_p$ is the treatment prescribed by the network instance $x_t$, corresponding to the lowest predicted outcome. The P-ReLU network is learned via gradient descent by minimizing equation \eqref{eq:obj}.

\subsection{Optimal Policy}

It is well-known  that  given a ReLU network, once an activation pattern on the hidden layers of the network is fixed, the network degenerates into a linear model. Feasible sets, coming from all feasible activation patterns, partition $\mathcal{X}$ into a finite number of polyhedra such that $\mathcal{P}_o \cap \mathcal{P}_{o'} = \emptyset, \ \forall o\neq o'$, and $\cup_{o} \mathcal{P}_o = \mathcal{X}$ \cite{serra2018bounding,lee2019towards}.

Our first theorem extends this result to the prescriptive setting, where the proposed architecture further refines disjoint polyhedra through the output layer to form clusters of instances that are being prescribed a same treatment. 

\begin{theorem}
P-ReLU network partitions the input space $\mathcal{X}$ into disjoint treatment polyhedra, where all instances that belong to the same polyhedron receive the same treatment.\label{thm:partition}
\end{theorem}

The proof is deferred to the Appendix. We present a toy example to illustrate the  partition scheme in Figure \ref{fig:toy_example}.

\noindent\textbf{Example 1: }Consider a setting with two features $x_1,x_2$, where $\mathcal{X}=[0,1]^2$, and a trained one-layer P-ReLU network with two hidden neurons and $K=2$ output nodes (treatments), as depicted in Figure \ref{fig:toy_example} (Left). We have $x^1 = (x_1-x_2,x_1+x_2-0.5)$ and $z^1= (\max\{x_1-x_2,0\},\max\{x_1+x_2-0.5,0\})$. By enumerating all possible activation patterns for the hidden layer, we obtain four convex polytopes, $\mathcal{P}_1, \mathcal{P}_2, \mathcal{P}_3 $ and $\mathcal{P}_4$, that partition the input space $\mathcal{X}$ as shown in Figure \ref{fig:toy_example} (Middle). 
Let's focus on finding the optimal policy for region $\mathcal{P}_1=\{(x_1,x_2)\in \mathcal{X}:x_1- x_2\geq0,x_1+x_2-0.5\geq0\}$. For $x \in \mathcal{P}_1$, 
$y_0  =1.5x_1 +2.5x_2-1$ and $y_1  =-0.5x_1-1.5x_2 +0.5$. Thus, we assign treatment $p=0$ to all $x \in \mathcal{F}_0 := \{x \in \mathcal{P}_1:y_0 <y_1\}$ and treatment $p=1$ to all $x \in \mathcal{F}_1 := \{x \in \mathcal{P}_1:y_0 \geq y_1\} = \mathcal{P}_1 \backslash \mathcal{F}_0$. Partitions $\mathcal{F}_0$ and $\mathcal{F}_1$, shown in Figure \ref{fig:toy_example} (Right), further divide $\mathcal{P}_1$ into two disjoint convex polytopes, where all instances within each region are prescribed the same treatment. By repeating this procedure for every $\mathcal{P}_i$, we can retrieve the final partitions of the input space, with their corresponding prescribed policies. 

Most existing literature on prescriptive trees \cite{kallus2017recursive, bertsimas2019optimal, biggs2021model,subramanian2022constrained} focus on trees with univariate splits, i.e.,  decision boundaries are generated by an axis-aligned function. 
With a P-ReLU network,  each partition of the input space  can be described by a finite number of hyperplane splits, resembling the behavior of a multivariate split tree, also known as an oblique tree. We  formally state this observation in the following theorem.

\begin{theorem}\label{thm:oblique}
A P-ReLU network can be expressed as an equivalent oblique prescriptive tree. 
\end{theorem}

To prove this result, the main idea is to convert the P-ReLU network into an equivalent locally constant network, where the output is a piecewise constant function of the input.
The conversion is achieved by replacing  the output layer of a trained P-ReLU network with $K(K-1)/2$ ReLU neurons to model the pairwise differences between the $K$ predicted treatment outcomes. Lastly, we invoke the result from \citet{lee2019oblique} which showed the equivalence between locally constant networks and decision trees with hyperplane splits and complete the proof. It is critical to note that with observational data, we only have access to partial feedback on the actions taken, as opposed to full labels as in the standard supervised learning setting described in \citet{lee2019oblique}. Furthermore, in the context of classification and regression tasks studied in \citet{lee2019oblique}, training locally constant networks requires solving a time-consuming dynamic programming procedure (or approximations of it). In the prescriptive setting, given the proposed network architecture, locally constant networks can be learned efficiently using gradient descent.

Previous studies in the field of prescriptive trees \cite{ bertsimas2019optimal} have acknowledged the potential benefits of utilizing hyperplane splits. However, utilizing such splits in trees has been hindered by computational tractability \cite{bertsimas2019optimal,inductive1, norouzi2015,carreira2018alternating}. 
Our approach provides an alternative method for training oblique prescriptive trees: first train (a potentially large) and expressive P-ReLU network via gradient descent; next, transform the trained network into a prescriptive tree. This procedure leads to a model with the expressiveness of a neural network, efficient computational training performance, as well as desirable properties associated with trees such as interpretability.

\begin{remark}
    The proof of Theorem \ref{thm:oblique} implies that we can train a P-ReLU with $N+K$ neurons, and then convert it to an oblique prescriptive tree of depth $N+K(K-1)/2$. 
\end{remark}

In our experiments, we observe that prescriptive trees extracted from P-ReLU networks are far more compact, after excluding neurons that are never activated. As pointed out in \citet{hanin2019deep}, even though  the theoretical maximum number of partitions increases exponentially with the depth of the ReLU network, in practice, the actual number is significantly smaller. In Section \ref{app:scalability} of the Appendix, we investigate how the number of final partitions and prescription accuracy change with respect to  the complexity of P-ReLU networks  on a real-world dataset. We observe that even the most complex P-ReLU networks generate a relatively modest number of partitions.

It is worth noting that one of the key advantages of P-ReLU is that it can achieve a \emph{balance} between strong prescriptive performance and interpretability. In scenarios where superior performance is required, users can efficiently train a large P-ReLU network (in terms of parameters) for enhanced prescriptive accuracy. Conversely, in scenarios where interpretability holds paramount importance, users can opt to train a small, sparse P-ReLU network. It can then be converted into a (shallow) interpretable tree, which is still more accurate than existing tree-based prescriptive models, as we will demonstrate numerically in Section~\ref{sec:experiment}.

\section{Constrained Prescription}\label{sec:constraints}

A common yet critical requirement for prescription tasks in practice is to incorporate constraints. For example, in the medical domain, oftentimes it is necessary to enforce a patient's dose of medication to be below a certain limit, given their vital signs. Incorporating constraints is difficult for many tree-based prescriptive algorithms as they rely on recursive partitioning of the data, making it challenging to impose constraints across multiple branches of a tree.  In what follows, we describe how to incorporate constraints in a P-ReLU  with minimal modification to the original network.

We consider prescriptive constraints of the form,
\begin{align}
    \emph{if $Ax > b$ then assign treatment that belongs  in $\mathcal{T}$}, \label{eq:con}
\end{align}
 where $ A\in \mathbb{R}^{c\times d}, \mathcal{T}\subset [K]$. We can explicitly impose the constraints by using $c+1$ extra neurons in the P-ReLU architecture. Specifically, for each linear constraint $a_i^Tx > b_i, i \in \{1,\dots,c\},$ where $a_i$ is the $i$-th row of $A$ and $b_i$ the $i$-th value of $b$,  we define a neuron with weight vector $a_i \in \mathbb{R}^d$ and bias term $b_i \in \mathbb{R}$, that takes as input $x$ and outputs the value $z^{con}_i(x) = a_i^Tx - b_i$. If $z^{con}_i(x)> 0$, then the $i$-th linear constraint is satisfied. To impose constraint \eqref{eq:con}, we need to have $z^{con}_i(x) > 0, \ \forall i \in \{1.\dots,c\}$. To verify the satisfiability, we define a final neuron with the indicator/binary step activation function $\sigma(x) = \mathbbm{1}[x>0]$, that takes as input the minimum of all $z^{con}_i(x)$ and outputs the value $z^{con}(x) = \mathbbm{1}[\{\min\{z_1^{con}(x),\dots, z_c^{con}(x)\}> 0]$. Since the weights of the neurons are predefined by \eqref{eq:con}, there is no need to train this part of the network, and thus we can use the non-differentiable indicator activation function. 
 
 Finally, to impose the prescriptive rule, we  multiply $z^{con}(x)$ with a large number $M$, and connect the output of this neuron to the output neurons of the treatments in $[K]\setminus \mathcal{T}$. If $z^{con}(x) = 0$, then $Ax>b$ is not satisfied and the output of the model is not affected. If $z^{con}(x) = 1 $, then $Ax>b$ is satisfied, and a large value $M$ is added to the output neurons that correspond to the treatments to be excluded. Adding a large value to these neurons ensures that the P-ReLU does not select the specific treatments, as the network assigns the treatment with the lowest predicted outcome to each instance. The idea of using a large value $M$ to impose constraints has been studied widely (under the name of big-$M$ constraint) in the field of linear optimization and optimal machine learning \cite{bertsimas1997introduction,bertsimas2019machine}. We now demonstrate how constraints can be incorporated into a P-ReLU network with an example.

\begin{figure}[ht]
\centering
\includegraphics[width=.5\linewidth]{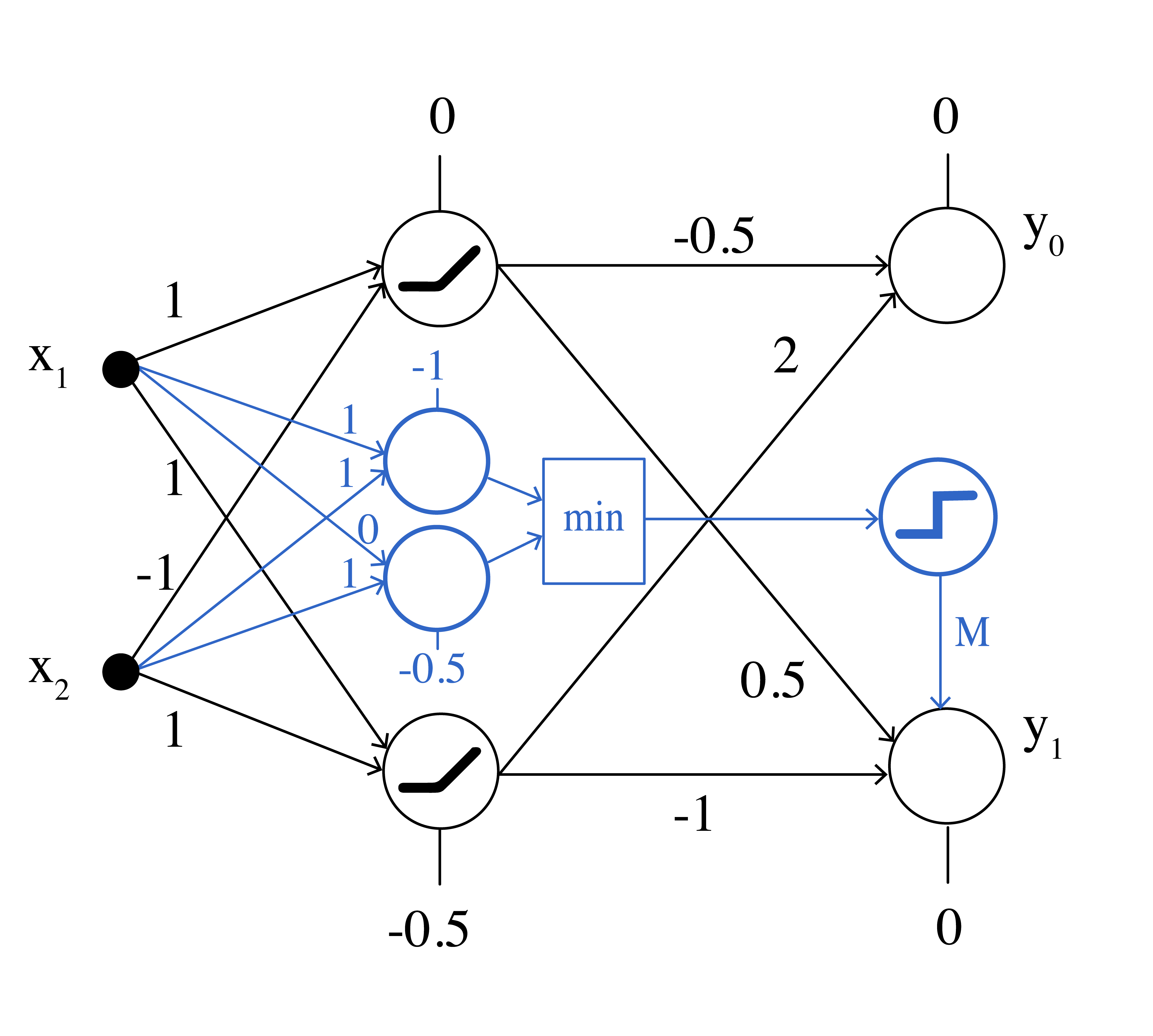}\hfill
\includegraphics[width=.5\linewidth]{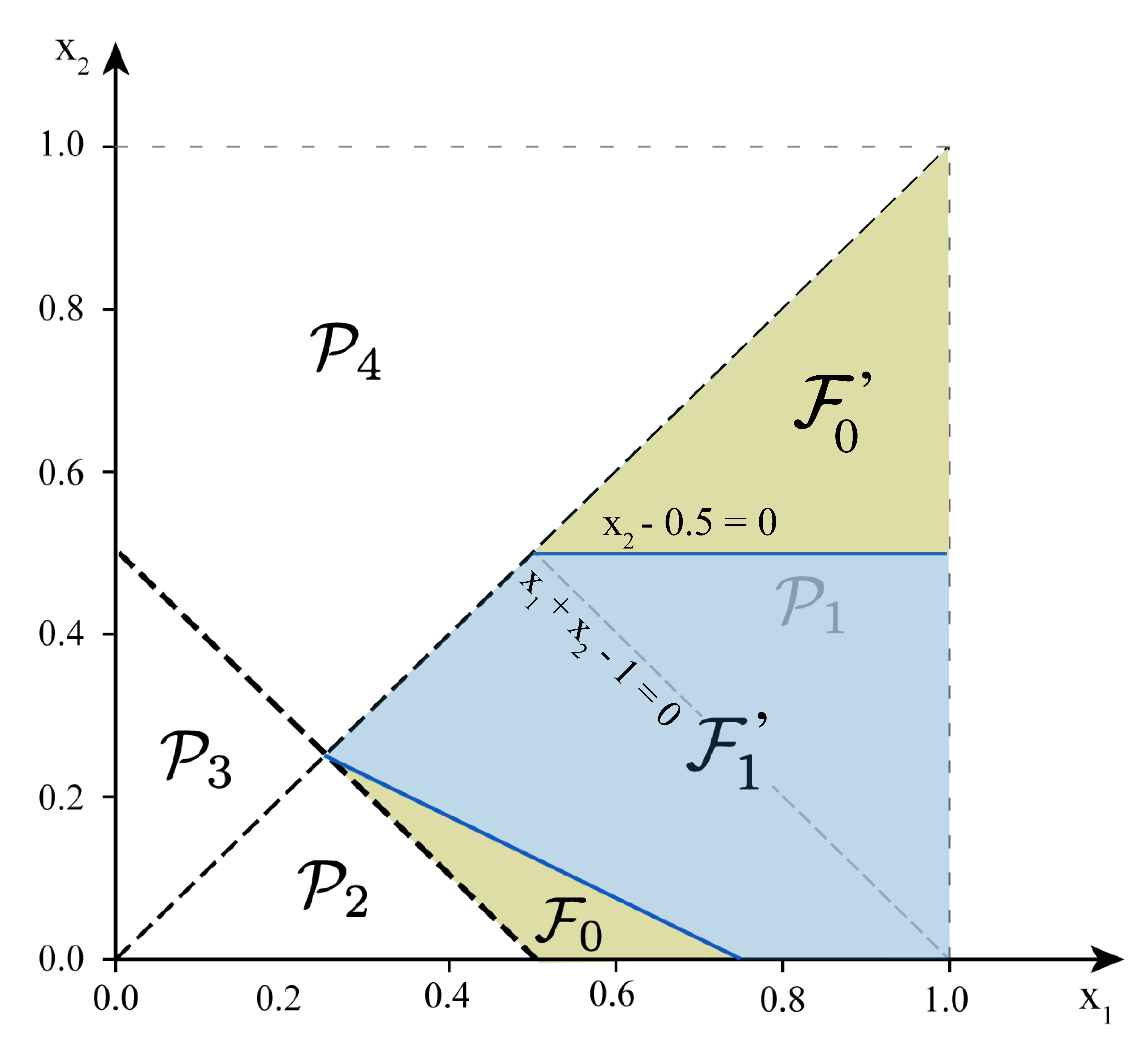}\hfill
\caption{An example that demonstrates a constrained P-ReLU network.
(Left) The newly added neurons that explicitly incorporate the prescription constraints. (Right) The final partition of $\mathcal{P}_1$ into the treatment regions $\mathcal{F}_0,\mathcal{F}^{'}_0$ and $\mathcal{F}^{'}_1$.}
\label{fig:toy_example_2}
\end{figure}

\noindent\textbf{Example 2: } We consider the P-ReLU of Figure \ref{fig:toy_example} (Left) with the addition of the rule \emph{if $x_1+x_2>1$ and $x_2>0.5$ then $T=\{0\}$}, and with big-$M$ value of $1,000$. In this case, we have $A = \begin{bmatrix}
    1 & 1\\
    0 & 1 
\end{bmatrix}$ and $b =\begin{bmatrix}
    1 \\
    0.5
\end{bmatrix}$. The constrained P-ReLU network is depicted in Figure \ref{fig:toy_example_2} (Left). In blue color, we observe the additionally added neurons that incorporate explicitly the prescription constraints. We focus again on $\mathcal{P}_1$, where now $y_0 = 1.5x_1 +2.5x_2-1$ and $y_1 = -0.5x_1-1.5x_2 +0.5 + M\cdot \mathbbm{1}[\{\min\{x_1+x_2-1,x_2-0.5\}> 0] $. For every  $x_2>0.5$ and $x_1 > 1-0.5=0.5$, we have that $y_1>y_0$ since $ M > 2x_1 +4x_2 -1.5$.  Therefore, the constrained P-ReLU prescribes treatment $0$ to all instances in $\mathcal{F}^{'}_0 = \{x \in \mathcal{P}_1: x_1 > 0.5, x_2 >0.5 \}$. Similarly, for treatment $1$, we obtain $\mathcal{F}^{'}_1$. This partition of $\mathcal{P}_1$ is depicted in \ref{fig:toy_example_2} (Right). Specifically, the initial $\mathcal{F}_1$ described in Figure \ref{fig:toy_example} (Right) is further partitioned into $\mathcal{F}^{'}_0$ and $\mathcal{F}^{'}_1$.

The previous example shows that a constrained P-ReLU still partitions the input space into disjoint convex polyhedra, as an oblique tree. 
\begin{corollary}\label{col:1}
Given a finite set of $C$ constraints $\{(A^{(1)},b^{(1)},\mathcal{T}^{(1)}),\dots,(A^{(C)},b^{(C)},\mathcal{T}^{(C)})\}$, the constrained P-ReLU network with $C+N+K$ neurons can be expressed as an equivalent oblique prescriptive tree of depth $C+N+K(K-1)/2$.
\end{corollary}

In Section \ref{app:nonlinear} of the Appendix, we discuss how this framework can be extended to incorporate some non-linear constraints, while in Section \ref{app:bigM}, we examine the selection of $M$ and the stability of training of the constrained P-ReLU model.

\section{Experiments}\label{sec:experiment}

\subsection{Benchmarks} \label{sec:benchmarks}

We test the P-ReLU model against the personalization tree (PT) and personalization forest (PF) \cite{kallus2017recursive}, optimal prescriptive trees (OPT) \cite{bertsimas2019optimal}, the causal tree (CT) \cite{athey2016recursive} and causal forests (CF) \cite{wager2018estimation} as well as the regress and compare (R\&C) approach with random forest (RF) and linear regression (LR).

Causal trees and forests are designed to choose between binary treatments by comparing $\mathbb{E}[Y | X = x, P = 1] - \mathbb{E} [Y | X = x, P = 0] $ to zero. To adapt CT and CF for multiple treatments with $K>2$, we incorporate the one-vs-all (1vA) and one-vs-one (1v1) schemes, similar 
to the procedures used in \citealp{kallus2017recursive}. We describe the schemes in detail in Section \ref{app:CTF} of the Appendix. It is important to note that the causal tree constructed for multi-treatments is no longer as interpretable as a single tree, given that there is no guarantee that the tree partitions associated with each treatment will align, resulting in a loss of the recursive tree structure and significantly more complex policies.

To train the causal tree and forest, we use the implementation from \citealp{econml}, using default parameters.  For training the personalization tree and as well as the personalization forest with 10 estimators, we use the python implementation from \citealp{PT_python}. We perform hyperparameter tuning via a grid search on the maximum tree depth $\Delta \in \{5,\infty\}$ and the minimum sample per leaf $n_{min-leaf} \in \{10,20\}$. We train OPTs with coordinate descent using the official implementation from \citet{InterpretableAI}. We perform hyperparameter tuning via the build-in grid search function on the maximum tree depth $\Delta \in \{5,\infty\}$ and the minimum sample per leaf $n_{min-leaf} \in \{10,20\}$. We test all four variants of OPTs, i.e. i) axis-aligned tree with constant predictions in its leaf for the treatment outcome, ii) axis-aligned splits with linear predictors, iii) hyperplane-splits with constant predictions, and iv) hyperplane-splits with linear predictors in its leaf. The last variant was excluded due to its prohibitive run time (more details later). For the random forest and linear regression used in the R\&C approach, we use the scikit-learn \cite{pedregosa2011scikit} with the default parameters. We report the best-performing PT, PF and OPT for each experiment, as well as the best-performing variant of CT and CF for each experiment with $K>2$.

\subsection{Simulated Data}\label{sec:simulated}

Simulated datasets allow us to accurately evaluate the counterfactual outcomes associated with changing treatments, and calculate the resulting outcome from a policy. Our experimental setup is adapted from \citealp{powers2018some} and \citealp{bertsimas2019optimal}. We generate samples with features  $x \in \mathbb{R}^{20}$. All samples are independent and identically distributed, with odd-numbered features coming from a standard Gaussian distribution, and even-numbered features coming from a Bernoulli distribution with probability 0.5.

For each experiment, we define a base function that gives the baseline outcome for each observation and an effect function that models the impact of the treatment applied. Both the baseline and the effect are functions of the features and are centered and scaled to have zero mean and unit variance. For the experiments with $K=2$ treatments, the outcome function is defined as $$y_p(x) = \mathrm{Base}(x) + (p-\frac{1}{2})\mathrm{Effect}(x),$$ where $p=0,1$. For experiments with $K = 3 $ treatments, the response is given by $$y_p(x) = \mathrm{Base}(x) + p(2-p)\mathrm{Effect}_1(x) +\frac{p(p-1)}{2}\mathrm{Effect}_2(x), $$ where $p=0,1,2$. In the training set, we add standard Gaussian noise with $\sigma^2 = 1$  to the outcomes that correspond to the selected treatment.

To simulate observational studies, we   assign treatments based on some propensity functions. More specifically, for $K=2$, we assign treatment $1$ with probability $\mathbb{P}[P=1|X=x] = \frac{e^{y_0(x)}}{1+e^{y_0(x)}}$. For $K=3$, we assign treatment $0$ with probability $\mathbb{P}[P=0|X=x] = \frac{1}{1+e^{y_0(x)}}$  and the other two treatments with probability $(1-\mathbb{P}[P=0|X=x])/2$.

The following functions are used to model the baseline and/or the effect functions. The intent is to capture a wide variety of functional forms including univariate and multivariate, additive and interactive, and piecewise constant, linear, and quadratic. 

\begin{itemize}
\item $f_1(x) = 5\mathbbm{1}[x_1>1]-5$,
\item $f_2(x) = 4\mathbbm{1}[x_1>1]\mathbbm{1}[x_3>0] + 4\mathbbm{1}[x_5>1]\mathbbm{1}[x_7>0] +2x_8x_9$,
\item $f_3(x)=0.5(x_1^2+x_2+x_3^2+x_4+x_5^2+x_6+x_7^2+x_8+x_9^2-11)$, 
\item $f_4(x) = x_2x_4x_6+2x_2x_4(1-x_6)+3x_2(1-x_4)x_6 +4x_2(1-x_4)(1-x_6)+5(1-x_2)x_4x_6+6(1-x_2)x_4(1-x_6)+7(1-x_2)(1-x_4)x_6 +8(1-x_2)(1-x_4)(1-x_6)$.
\end{itemize}

We generate six datasets consisting of different combinations of the baseline and effect functions as shown in Table~\ref{tab:specifications}. For each dataset, we create $10,000$ training samples and $5,000$ testing samples.

\begin{table}[ht]
\small
\centering
 \caption{Specifications for the simulated datasets.}
\label{tab:specifications}
\begin{tabular}{ccc}
\toprule 
    Dataset & Base & Effect  \\ \midrule
    1 & $f_1(x)$    &  $f_2(x)$ \\
    2 & $f_4(x)$    &  $f_2(x)$  \\
    3 & $f_3(x)$    &  $f_4(x)$   \\
    4 & $f_1(x)$     & $f_3(x)$ \\
    5 & $f_2(x)$    &  $f_1(x),f_3(x)$  \\
    6 & $f_2(x)$    &  $f_3(x),f_4(x)$  \\
\bottomrule
\end{tabular} 
\end{table}

For our proposed method, we consider a five-layer P-ReLU network with $100$ neurons per hidden layer. We train the model using Adam optimizer \cite{kingma2014adam} with learning rate equal to $10^{-3}$ for $20$ epochs, batch size equal to $64$, and $\mu = 10^{-4}$. We compare our approach (P-ReLU) with the benchmark algorithms described earlier. We run 10 independent simulations for each dataset and report the mean and standard deviation of prescription accuracy which measures the proportion of the test set prescribed with the correct treatment in Table \ref{tab:syn_results_total}.

\begin{table*}[ht]
\small
\centering
\caption{Mean prescription accuracy (\%) for the synthetic datasets.}
\label{tab:syn_results_total}
\begin{tabular}{ccccccc}
\toprule 
    Algorithm & Dataset 1 & Dataset 2 & Dataset 3 & Dataset 4 & Dataset 5 & Dataset 6 \\
     \midrule
     R\&C RF	 &$76.03 \pm 3.55$ & $ 60.59	 \pm 0.70 $ &  $ 99.95 \pm 0.06 $ & $\textbf{93.71} \pm 0.31$ & $ 82.19	 \pm 1.35 $ &  $ 81.13 \pm 1.29 $ \\ 
     R\&C LR & $56.70 \pm 0.50$ & $ 57.33\pm  0.39	$ & $ \textbf{100} \pm 0.00 $ & $83.30 \pm 0.86$ & $ 66.58 \pm 1.47 $ &  $ 50.08 \pm 1.82 $\\ 
     CT  &  $53.56 \pm 1.81$ & $ 57.79 \pm 0.54$ & $ 90.85 \pm 7.35 $ & $43.79 \pm 3.35$ & $ 67.66 \pm 1.49 $ &  $ 75.41 \pm 2.28 $\\ 
     CF  & $50.80 \pm 1.02$ & $ 57.86 \pm 0.51$ &  $ 95.33 \pm 1.32 $ & $\textbf{93.37} \pm 0.29$ & $ 71.94 \pm 1.39 $ &  $ 72.56 \pm 1.53 $\\ 
     PT &  $51.60 \pm 1.12$ & $ 57.41 \pm 0.39$ &  $ \textbf{100} \pm 0.00 $ & $\textbf{93.25} \pm 0.29$ & $ 74.24 \pm 0.43$ &  $ 67.58 \pm 4.38 $\\
     PF & $51.43 \pm 1.03$ & $ 57.41 \pm 0.39 $ &  $ \textbf{100} \pm 0.00 $ & $\textbf{93.45} \pm 0.31$ & $ 74.26 \pm 0.37 $ &  $ 68.49 \pm 3.96 $\\
     OPT & $57.03 \pm 3.04$ & $60.37 \pm 2.24$ & $\textbf{100} \pm 0.00 $ & $\textbf{93.14} \pm 1.65 $& $65.38 \pm 0.93$ & $64.83 \pm 2.52$\\
     P-ReLU  &$\textbf{84.47} \pm 2.50$ & $\textbf{66.72} \pm 2.32 $ &  $\textbf{100} \pm 0.00 $ & $\textbf{93.75} \pm 1.18$&  $ \textbf{88.80} \pm 1.42 $ &   $ \textbf{87.08} \pm 2.14 $  \\
\bottomrule
\end{tabular} 

\end{table*}

We observe that the P-ReLU network demonstrates superior or comparable performance compared to competing prescriptive methods in all experiments. Overall, R\&C with random forest is the first runner-up, closely trailing behind P-ReLU. We also observe that the margin between P-ReLU and other methods widens when there are more treatments. Since our proposed method utilizes pooled data in learning the response functions, whereas many other competing methods (e.g., R\&C based approach, CT, PT, OPT) divide the data into specific treatment subpopulations, we believe P-ReLU will have more advantages in scenarios with many treatments. Moreover, in these sets of experiments, the outcome is a composite of quadratic and linear or piecewise linear treatment effects, which is more challenging for competing models to learn. For instance, OPTs assume constant or linear response functions at each leaf. On the other hand, P-ReLUs can in theory approximate any continuous function to model the counterfactuals. 

We also want to  point out that 
even though OPTs are trained using coordinate descent, scalability continues to be a challenge since its underlying mixed-integer formulation relies on a huge number of binary decision variables 
that is in the order of  $\mathcal{O}(2^Dn)$, where $D$ is the depth of the tree and $n$ is the sample size.
For OPT's most complex variant, which utilizes hyperplane-splits with linear predictors in its leaf, a single instance took more than 14 hours to complete. Thus, it was omitted from the experiments. 
A detailed discussion on OPTs and their runtimes as well as a comparison with P-ReLUs can be found in Section \ref{app:compare_opt} of the Appendix.

\subsection{Personalized Warfarin Dosing}

Warfarin is the most widely used oral anticoagulant agent worldwide according to the International Warfarin Pharmacogenetics Consortium. Finding the appropriate dose for a patient is difficult since it can vary by a factor of ten among patients and incorrect doses can contribute to severe adverse effects \cite{international2009estimation}. The current guideline is to start the patient at 35 mg per week, and then vary the dosage based on how the patient reacts \cite{jaffer2003practical}.

The dataset was collected and curated by Pharmacogenetics and Pharmacogenomics Knowledge Base and the International Warfarin Pharmacogenetics Consortium. One advantage of this dataset is that it gives access to counterfactuals as it contains the true stable dose found by physician-controlled experimentation for $5,701$ patients. After pre-processing the dataset by excluding missing values and transforming categorical ones, we obtain a  dataset of $4,416$ patients. The patient covariates include demographic information (sex, race, age, weight, height), diagnostic information (reason for treatment, e.g., deep vein thrombosis, etc.), and genetic information (presence of genotype polymorphisms of CYP2C9 and VKORC1). The correct stable therapeutic dose of warfarin is segmented into three dose groups: Low ($\leq 21 \text{ mg/week }$), Medium ($>21, <49 \text{ mg/week }$), and High ($\geq 49 \text{ mg/week }$), corresponding to $p=0,1$ and 2 respectively. 

As the dataset contains the correct treatment for each patient, one can develop a prediction model that treats the problem as a standard multi-class classification problem with full feedback which predicts the correct treatment given the covariates. Solving this classification problem gives an upper bound on the performance of prescriptive algorithms trained on observational data, as this is the best they could do if they had complete information.  We considered random forest (RF), support vector classifier(SVC), logistic regression (LogReg), kNN, and  XGBoost (XGB) classifier using scikit-learn. We perform a total of  $10$ runs and at each run, we tune the hyper-parameters of the classifiers by performing grid-search using cross-validation with 3 folds. The mean accuracy per classifier is reported in Table \ref{tab:war_clas} of the Appendix. We conclude that the best performing model here is the logistic regression classifier with mean accuracy  $\approx 69.11\%$.

To modify the dataset suitable as an observational study, we follow the  experimental setup described in \citet{kallus2017recursive}. We consider the treatment chosen based on body mass index (BMI) by the corresponding probabilities $\mathbb{P}[P=p|X=x]=\frac{1}{S}\exp\left(\frac{(p-1)(BMI-\mu)}{\sigma}\right),$ for $p=0,1,2$, where $\mu,\sigma $ are the mean and standard deviation of patients’ BMI and $S$ is the normalizing factor. We set the outcome $y_t(p)$ to be $1$ if the dose $p$ is incorrect for instance $t$ and $0$ otherwise. 

We implement a prescriptive ReLU net with five hidden layers and $100$ neurons per layer that we train using the parameters described in Section \ref{sec:simulated}. We perform a total of $10$ runs and we report the mean prescription accuracy of our models and the benchmarks in Table \ref{tab:war_results}.

\begin{table}[ht]
\centering
    \caption{Mean prescription accuracy (\%) of each prescriptive algorithm for the warfarin prescription problem.}
\label{tab:war_results}
\begin{tabular}{cc}
\toprule 
    Algorithm & Accuracy \\
     \midrule
     R\&C RF	 & $ 65.43 \pm 1.38 $\\ 
     R\&C LR & $ 66.68\pm  0.67$\\ 
     CT  & $ 54.87 \pm 0.68 $\\
     CF & $ 64.80 \pm 0.80$\\ 
     PT & $ 55.43 \pm 1.53$\\
     PF  & $ 56.03\pm 0.98 $\\
     OPT & $65.04 \pm 1.63$ \\
     P-ReLU & $ \textbf{68.27} \pm 0.76$\\
\bottomrule
\end{tabular} 

\end{table}

We observe that the highest mean prescription accuracy $\approx 68.27\%$, is achieved by the five-layer prescriptive ReLU neural network. The accuracy of the P-ReLU network trained on observational data, is a mere  $\approx1\%$ less than the highest accuracy for the multi-class classification problem with full information, as shown in Table \ref{tab:war_clas}.   

To generate an interpretable prescription policy we train a one-layer P-ReLU network with $5$ neurons. To enhance interpretability, we perform sparse training using weight pruning where we keep $1$ weight per hidden neuron.  Specifically, the sparse training is performed at each training iteration where for each neuron we keep only the largest in terms of absolute value weight and we set the rest of the weights to zero. The bias terms of the neurons are not affected by this procedure. This technique is known as targeted dropout/magnitude-based pruning \cite{targeted-dropout}.  The mean prescription accuracy for this ``sparse'' model is $64.48 \%$. While this simpler single-layer P-ReLU is less accurate than its more expressive five-layer counterpart, its performance is significantly higher than the single-tree-based methods (i.e., PT, CT, and OPT) which maintain interpretability. Note that in this experiment, the best-performing OPT reported in Table \ref{tab:war_results} (with mean prescription accuracy of $65.04\% \pm 1.63\%$)
corresponds to the one with hyperplane splits and constant predictions. Meanwhile, the mean prescription accuracy of the more interpretable OPT with axis-aligned splits and constant predictions is lower, at $63.57\% \pm 1.23\%$.
 
\begin{figure}[ht]
\centering
\includegraphics[scale=0.33]{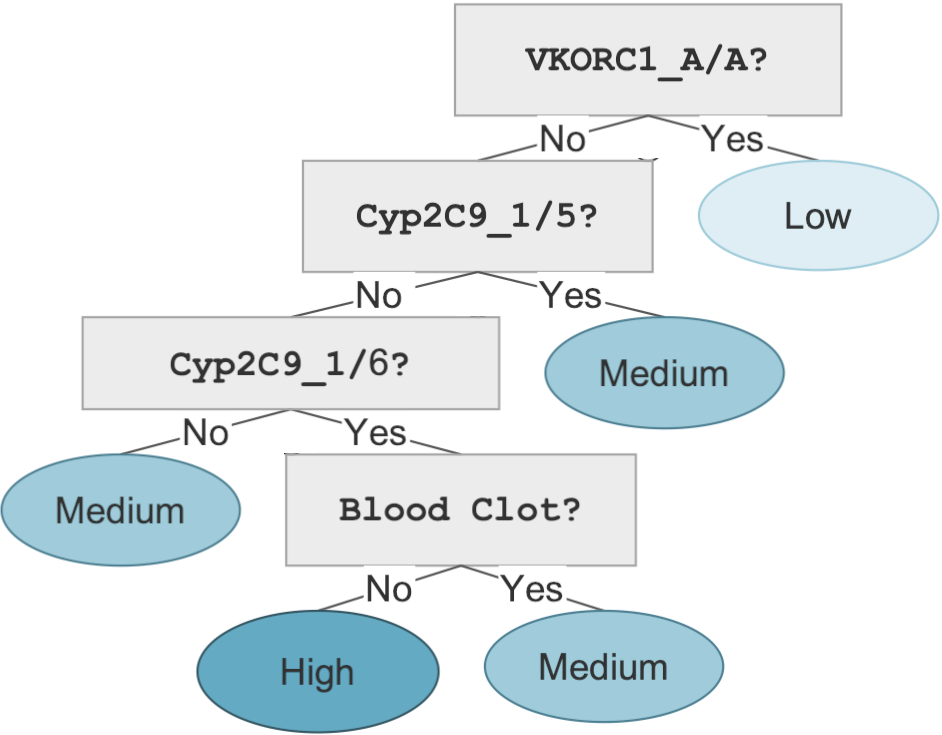}
\caption{The equivalent prescriptive tree of the 1-layer P-ReLU network for the unconstrained warfarin dosing problem.}

\label{fig:warfarin_tree_uncon}
\end{figure}

By following the procedure described in the proof of Theorem \ref{thm:oblique}, we can convert the network into a simple prescriptive tree presented in Figure \ref{fig:warfarin_tree_uncon}. Note that before we convert the P-ReLU into a tree,  we remove neurons that are always deactivated. To do so, we perform a feed-forward pass using the training data and we keep track of whether a neuron was activated or not for each training sample. If a neuron is deactivated for all training samples we can completely remove it from the architecture. Also, if a neuron is activated for all training samples we can also remove it and adjust the corresponding weights of the next layer. By performing this step, we reduce the total number of neurons in the P-ReLU network, which leads to smaller resulting trees, as meaningless splits are not performed, and faster inference times.

We observe that the proposed P-ReLU splits the input space into $5$ partitions. Each partition corresponds to a leaf of the tree and can be described by the splits that lead to the specific leaf. The most prevalent features in the tree are whether VKORC1 and CYP2C9 genotypes are present and whether the patient had a pulmonary embolism (PE, i.e. when a blood clot has traveled to the lung), which are known to be strongly associated with warfarin dosage requirements \cite{li2006polymorphisms}.

\subsection{Constrained  Warfarin Dosing}

Studies suggest that BMI is positively correlated with  warfarin dosage \cite{mueller2014warfarin}, i.e., the average weekly warfarin dosage can be approximated by $12.34 + 0.69 \cdot BMI$. Based on this result, we design the following constraint:  \emph{if $BMI>30$, then treatment in $\mathcal{T}=\{Medium, High\}$}. Note that the unconstrained policy in Figure \ref{fig:warfarin_tree_uncon}, does not take into account BMI, and it can be verified that some instances indeed violate this constraint. 

\begin{figure}[ht]
\centering
\includegraphics[width=\linewidth]{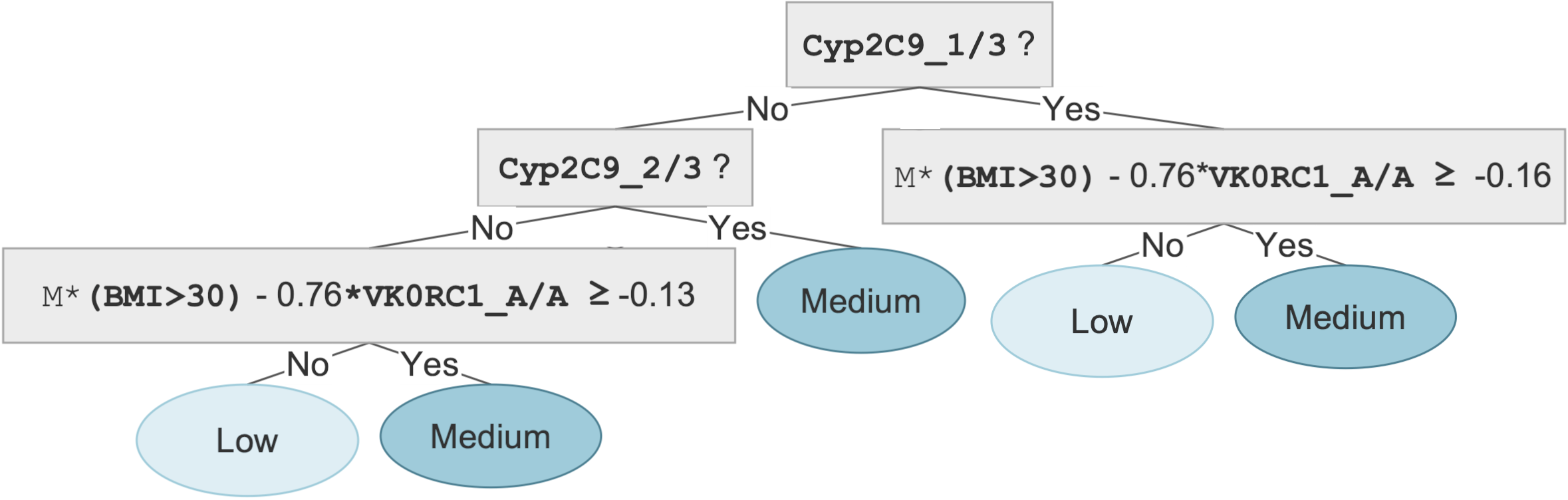}
\caption{The equivalent prescriptive tree of the constrained 1-layer P-ReLU network.
}

\label{fig:warfarin_tree_con}
\end{figure}

We utilize the same P-ReLU setup described for the unconstrained setting and apply the methodology described in Section \ref{sec:constraints} with $M=1,000$ 
to incorporate the constraint. The 
resulting tree which achieves the prescription accuracy of  $64.33\%$ is depicted in Figure \ref{fig:warfarin_tree_con}.

The big-$M$ appears in the hyperplane splits and explicitly enforces the constraint associated with BMI. To see this, consider the first right child node whose splitting criterion is $M*(BMI>30)-0.76*VKORC1\_A/A \geq -0.16$. Since all the features are binary, when $BMI >30$,  the criterion is True because of the big-$M$ and  $Medium$ dosage is prescribed. On the other hand, when $BMI \leq 30$, the optimal policy will choose between $Low$ or $Medium$ depending on whether VKORC1\_A/A genotype exists. 

\section{Conclusion}
Our work adds to a burgeoning body of research, i.e., policy optimization from observational data for decision-making. We propose an integrated approach, namely P-ReLU neural networks, that simultaneously determines the optimal policy and estimates the counterfactuals from data. We highlight the flexibility of the proposed method as constraints which are often critical for practical applications can be easily incorporated. Finally, by transforming a trained  P-ReLU into an oblique tree, we illustrate the interpretability of learned policy.  Future research could investigate extending this approach to other types of neural networks and explore the trade-offs between interpretability and performance in policy learning scenarios. Additionally, it could be interesting to incorporate other types of structural constraints (such as monotonicity and convexity) into the P-ReLU model.

\bibliography{example_paper}
\bibliographystyle{icml2023}

\newpage
\appendix
\onecolumn
\section{Proof of Theorem \ref{thm:partition}}

\textbf{Theorem 3.1} \textit{P-ReLU network partitions the input space $\mathcal{X}$ into disjoint treatment polyhedra, where all instances that belong to the same polyhedron receive the same treatment.}

\begin{proof}
    Let $f_{\theta}(\cdot)$ be an already trained P-ReLU network and $o$ a feasible activation pattern. We denote as $\mathcal{P}_o \subset \mathcal{X}$ the corresponding to the activation pattern $o$ convex polyhedron. The network assigns to each instance the treatment with the lowest predicted response. Assuming $K$ treatments and the fixed activation pattern $o$, the network degenerates to $K$ linear models over $\mathcal{P}_o$ that we denote as $f_{\theta,o}(\cdot)$. To determine the prescription of a specific treatment to an instance, we need to solve a system of $K-1$ linear inequalities 
    over $\mathcal{P}_o$. We define the prescription region for a treatment $i \in [K]$ as the solution region for the system of linear inequalities defined as $\mathcal{F}_{o,i} := \mathcal{P}_o\cap \{x: f_{\theta,o}(x)_i < f_{\theta,o}(x)_j, \ \forall j \in [K], i\neq j \}$ (we break ties arbitrarily, but consistently). Therefore, every $x \in \mathcal{P}_o$ that also belongs in $\mathcal{F}_{o,i}$ receives treatment $i$. By definition $\mathcal{F}_{o,i} \cap \mathcal{F}_{o,j} = \emptyset, \ \forall i\neq j$, and $\cup_{i \in [K]} \mathcal{F}_{o,i} = \mathcal{P}_o$.  
    
    By enumerating all possible feasible activation patterns, we observe that $\cup_{o}(\cup_{i \in [K]} \mathcal{F}_{o,i}) = \cup_{o}\mathcal{P}_o  = \mathcal{X}$. Therefore, the proposed network partitions the input space into disjoint convex polyhedra that prescribe the same treatment to instances that belong to the same partition.
\end{proof}

\section{Proof of Theorem \ref{thm:oblique}}

We first state the formal definition of an oblique tree  \cite{lee2019oblique}.

\begin{definition} 
The class of oblique trees contains any functions that can be defined procedurally, with depth $D \in \mathbb{Z}_{>0}$, for $x \in \mathbb{R}^d$:
\begin{enumerate}
    \item $r_1 := \mathbbm{1}[\omega_{\emptyset}^Tx + \beta_{\emptyset}\geq 0]$, where $\omega_{\emptyset} \in \mathbb{R}^d$ and $\beta_{\emptyset} \in \mathbb{R}$ denote the weight and the bias of the hyperplane split at the root node.
    \item  For every $i \in \{2,\dots,D\}$, $r_i := \mathbbm{1}[\omega_{r_{1:i-1}}^Tx + \beta_{r_{1:i-1}}\geq 0]$, where $\omega_{r_{1:i-1}} \in \mathbb{R}^d$ and $\beta_{r_{1:i-1}} \in \mathbb{R}$ denote the weight and the bias of the  hyperplane split of the decision node after the decision pattern $r_{1:i-1}$.
    \item $v: \{0,1\}^D \to \mathbb{R}$ outputs the leaf value $v(r_{1:D})$ associated with the decision pattern $r_{1:D}$.
\end{enumerate}
\end{definition}
We now prove formally that each P-ReLU can be expressed by an equivalent oblique tree.

\textbf{Theorem 3.2} \textit{A P-ReLU network can be expressed as an equivalent oblique prescriptive tree. }

\begin{proof}
We first convert the already trained P-ReLU network  $f_{\theta}$ into a locally constant neural network. Locally constant neural networks, which can also be represented by ReLU networks, output a constant value for each activation pattern $o$. More formally, for a given activation pattern $o(x)$, the output of the network is $g(o(x))$ with $g: \{0, 1\}^N \to \mathbb{R}$, where $N$ is the total number of the hidden neurons of the network (i.e., $N=\sum_{i=1}^L N_i$ with $L$ being the number of hidden layers of the locally constant network). In general, function $g$ can be seen as a lookup table.

We convert the P-ReLU network $f_{\theta}$ into a locally constant network by changing the output layer as follows. By fixing the $L$ hidden layers of $f_{\theta}$, we replace the $K$ neurons of the output layer $L+1$ with $\frac{K(K-1)}{2}$ ReLU neurons. Each newly added neuron models the difference between the estimated treatment outcomes. Formally, the output of the neuron $(i,j)$ after the ReLU activation, with $i \in \{1,\dots,K-1\}, j \in \{i+1, \dots, K\}$, is $z^{L+1}_{(i,j)} = \max\{  (W^{L+1}x^L+b^{L+1})_i - (W^{L+1}x^L+b^{L+1})_j, 0\} = \max\{f_{\theta}(x)_i-f_{\theta}(x)_j,0\} $. Intuitively, each neuron $(i,j)$ outputs a positive value if the predicted outcome for treatment $i$ is greater than the predicted outcome for treatment $j$ and zero otherwise. Therefore, if we define $g: \{0, 1\}^{N+K(K-1)/2} \to \{0,\dots, K-1\}$ to output treatment $0$ if $z^{L+1}_{(0,1)}=z^{L+1}_{(0,2)}=\dots=z^{L+1}_{(0,K-1)}=0$, $1$ if $z^{L+1}_{(0,1)}=1$ and $z^{L+1}_{(1,2)}=z^{L+1}_{(1,3)}=\dots=z^{L+1}_{(1,K-1)}=0$ and so on, independent of the value of the rest of the activations of the previous $L$ layers of the network, we obtain a locally constant network that is equivalent to $f_{\theta}$.

Then, we can invoke the result stated in the proof of Theorem 3 of \cite{lee2019oblique} and we can re-write the locally constant network to have $1$ neuron per layer, by expanding any layer with $N_i>1$ neurons to be $N_i$ different layers, such that they do not have intra-connections. Then, following the notation of a network with one neuron per layer, we can define the following equivalent oblique decision tree with depth $D=N+\frac{K(K-1)}{2}$:
\begin{enumerate}
     \item $r_1 := o_1^1(x) = \mathbbm{1}[\omega_{\emptyset}^Tx + \beta_{\emptyset}\geq 0]$, where $\omega_{\emptyset} = W^1_{1,:}$ and $\beta_{\emptyset} =b^1_1$ .
    \item  For every $i \in \{2,\dots,D\}$, $r_i := \mathbbm{1}[\omega_{r_{1:i-1}}^Tx + \beta_{r_{1:i-1}}\geq 0]$, where $\omega_{r_{1:i-1}} = \nabla_x z^i_1$ and $\beta_{r_{1:i-1}} = z^i_1 - (\nabla_x z^i_1)^Tx$.  Please note that $r_i = o^i_1(x)$.
    \item $v=g$.
\end{enumerate}

To obtain a valid oblique prescription tree, $\omega_{r_{1:i-1}}$ and $\beta_{r_{1:i-1}}$ have to be unique for all inputs that yield the same activation pattern. Since $r_{1:i-1} = (o_1^1(x),\dots,o_1^{i-1}(x))$, $z^i_i$ is a fixed affine function given an activation pattern for the previous neurons. As a result, $z^i_1 - (\nabla_x z^i_1)^Tx$ and $(\nabla_x z^i_1)$ are fixed quantities for all $x$ that produce the same activation pattern. Finally, having $r_{1:D} = o^D(x)$ and $v=g$ concludes the proof.
\end{proof}

\section{Proof of Corollary \ref{col:1}}

\textbf{Corollary 4.1} \textit{Given a finite set of $C$ constraints $\{(A^{(1)},b^{(1)},\mathcal{T}^{(1)}),\dots,(A^{(C)},b^{(C)},\mathcal{T}^{(C)})\}$, the constrained P-ReLU network with $C+N+K$ neurons can be expressed as an equivalent oblique prescriptive tree of depth $C+N+K(K-1)/2$.}
\begin{proof}
As in the proof of Theorem \ref{thm:oblique}, we have to convert the already trained constrained P-ReLU network  $f^c_{\theta}$ into a locally constant neural network. At the first layer of the constrained P-ReLU network, we define $C$ new artificial neurons, whose output is given by $z^{(i)}(x) = \mathbbm{1}[\{ \min\{(a^{(i)}_1)^Tx + b^{(i)}_1,(a^{(i)}_2)^Tx + b^{(i)}_2, \dots\} > 0 \}$, $i =1,\dots,C$. Then, for each new neuron $i$, we create $L$ new ReLU neurons, one at each hidden layer, that propagate $z^{(i)}(x)$ to the output layer. These $L$ new neurons have all input weights equal to $0$, except the weight from the previously newly introduced neuron that is equal to $1$. Their biases are equal to $0$ as well. 
    
    In the final newly added neuron before the output layer, the weights between the newly added neuron and the output neurons $[K]\setminus \mathcal{T}^{(i)}$ are set to $M$, while all other weights are $0$. By transforming the output layer, as in the proof of Theorem \ref{thm:oblique}, we obtain that the output of the neuron $(i,j)$ after the ReLU activation, with $i \in \{1,\dots, K-1\}, j \in \{i+1, \dots, K\}$, is $z^{L+1}_{(i,j)} = \max\{  (W^{L+1}x^L+b^{L+1})_i + M\cdot \sum_{k=1}^Cz^{(k)}(x)\mathbbm{1}[i \in [K]\setminus \mathcal{T}^{(k)}]  - (W^{L+1}x^L+b^{L+1})_j - M\cdot \sum_{k=1}^Cz^{(k)}(x)\mathbbm{1}[j \in [K]\setminus \mathcal{T}^{(k)}], 0\} = \max\{f^c_{\theta}(x)_i-f^c_{\theta}(x)_j,0\} $, where $z^{(k)}(x)$ is the binary output of the neuron that corresponds to constraint $k, \ k \in\{1,\dots,C\}$. If we define $g: \{0, 1\}^{C+N+K(K-1)/2} \to \{0,\dots, K-1\}$ to output treatment $0$ if $z^{L+1}_{(0,1)}=z^{L+1}_{(0,2)}=\dots=z^{L+1}_{(0,K-1)}=0$, $1$ if $z^{L+1}_{(0,1)}=1$ and $z^{L+1}_{(1,2)}=z^{L+1}_{(1,3)}=\dots=z^{L+1}_{(1,K-1)}=0$ and so on, we obtain a locally piecewise constant network that is equivalent to $f^c_{\theta}$. The rest of the proof is as in Theorem \ref{thm:oblique} with the only difference that we define $r_i := z^{(i)}(x)$ for the neuron of the input layer that corresponds to the $i$-th constraint.
\end{proof}

\section{On the Number of Partitions of the P-ReLU network for the Personalized Warfarin Dosing Problem}\label{app:scalability}

\begin{figure}[ht]
\centering
\includegraphics[width=0.95\textwidth]{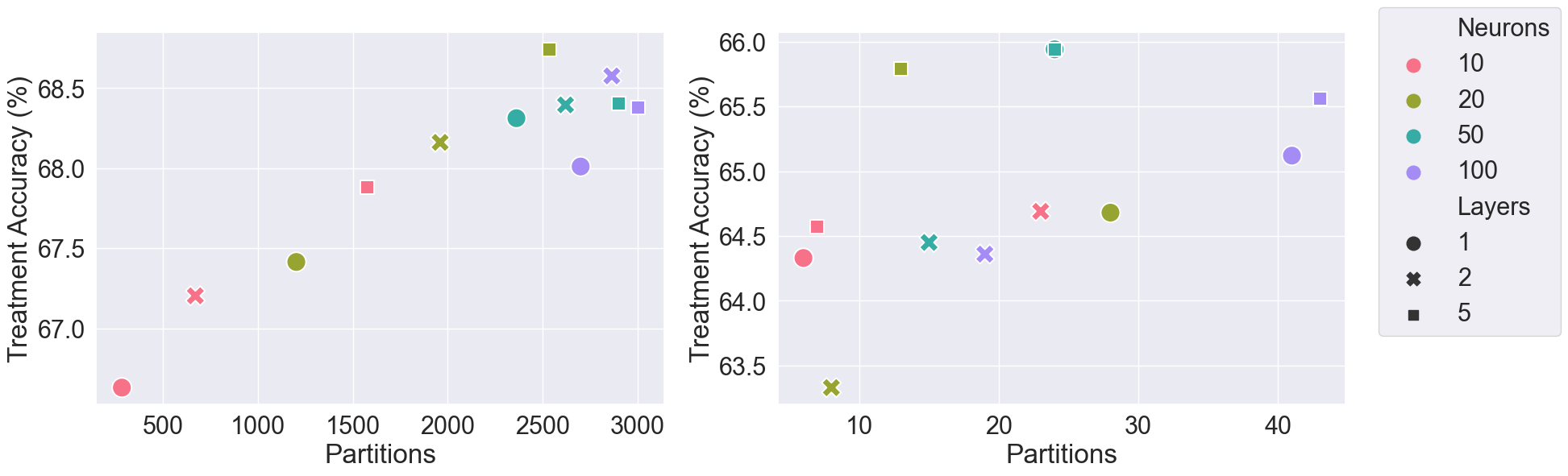}
\caption{Treatment accuracy with respect to the number of partitions and the network architecture (Left complete; Right sparse) for the personalized warfarin dosing problem.}
\label{fig:partitions}
\end{figure}

We investigate the relationship across the number of partitions (or activation patterns), the network architecture, and the treatment accuracy of the P-ReLU network for the personalized warfarin dosing problem. Specifically, we solve the problem by training multiple P-ReLU networks with $1,2$ and $5$ hidden layers, as well as, $5,10,20,50$ and $100$ neurons per hidden layer. We also perform the experiment with the sparse P-ReLU. In Figure \ref{fig:partitions}, we present the treatment accuracy with respect to the number of partitions for all examined network complexities.

We observe that even the most complex, non-sparse, P-ReLU network with $5$ hidden layers and $100$ neurons per layer ($500$ hidden neurons in total) partitions the input space $\mathcal{X}$ into $\approx 3,000$ regions/prescription rules. Even though in theory the maximum number of partitions/activation patterns for a ReLU network with $5$ hidden layers and $\nu=100$ neurons per layer and input dimension $40$  is of $O(\nu^{Ld})$ \cite{serra2018bounding}, this number is significantly smaller for this experiment.

\section{Extension to Non-Linear Prescription Constraints}\label{app:nonlinear}

The proposed approach on incorporating prescriptive constraints, described in Section \ref{sec:constraints}, focuses on linear prescriptive constraints of the form 
\begin{align*}
    \emph{if $Ax > b$ then assign treatment that belongs in $\mathcal{T}$}.
\end{align*}
It's worth noting that the proposed approach can also incorporate nonlinear constraints. This can be achieved by first transforming the original features and generating new, nonlinear ones at the input layer of the network, and then applying the rules as mentioned in Section \ref{sec:constraints}. For example, if the rule was in the form of \emph{if $(height)^2 + \log(weight) + (height)\cdot(weight) > b$ then prescribe treatment that belongs in $\mathcal{T}$}, we would first create the new features $(height)^2, \log(weight),(height)\cdot(weight)$ and then we would enforce the rule in the P-ReLU network as a typical linear prescriptive constraint.

\section{On the Value of $M$ and the Stability of Training for the Constrained P-ReLU}\label{app:bigM}

\begin{remark}
Assuming that (i) the input space $\mathcal{X}$ and (ii) the weights of the network are bounded, one can set $M$ to be the absolute value of the difference between an upper bound of the largest and a lower bound of the lowest value an output of the P-ReLU network can take. A simple upper bound can be found by setting each weight of the network to the largest possible absolute value and by performing a feed-forward pass with the largest possible absolute value of each feature as input. A lower bound can be retrieved accordingly. Both assumptions are mild and are achieved in practice by applying standard techniques to speed up training/avoid overfitting, such as normalizing the input data and weight normalization. In practice, the big-$M$ value of $1,000$ that we used in the experiment worked well without issues. \end{remark}
\vspace{0.4cm}

\begin{remark}
Assuming that there are no samples in the training dataset that violate the given prescriptive constraints, the stability of the training remains the same as in the unconstrained P-ReLU setting. This can be seen by observing both the prescriptive and the predictive parts of the objective. Regarding the prescriptive part of the objective described in equation \eqref{def_prescription}, given that P-ReLU always selects the treatment that corresponds to the output with the lowest value, the predicted outputs that incorporate the $M$ constant will not appear in the $\sum_{p \neq p_t}\hat{y}_t(p)\mathbbm{1}[\pi(x_t)=p]$ part. Regarding the predictive part of the objective described in equation \eqref{def_prediction}, under the aforementioned assumption, $\hat{y}_t(p_t)$ will never incorporate the $M$ constant, as $p_t$ is a feasible treatment according to the prescription constraints. Since both parts of the objective will not incorporate $M$, it is evident that will $M$ never appear in the loss during training. As a result, we will not come across numerical/stability issues during training due to the large value of $M$. The stated assumption is mild and can be achieved by removing the training samples that violate the prescription constraints from the dataset before training.\end{remark}

\section{On the Objective of the Proposed Loss Function}

The objective of the loss function defined in (\ref{def_loss}) can be rewritten as 
\begin{align*}
    \mu \cdot \sum_{t=1}^n \big(y_t \mathbbm{1}[\pi_{f_{\theta}}(x_t) = p_t] + \sum_{ p\neq p_t} f_{\theta}(x_t)_p  \mathbbm{1}[\pi_{f_{\theta}}(x_t) = p] \big)
    +(1-\mu) \cdot  \sum_{t=1}^n (y_t -  f_{\theta}(x_t)_{p_t})^2,  \numberthis \label{eq:obj_again}
\end{align*}
\noindent where $f_{\theta}(x_t)_p$ is the predicted outcome to treatment $p$ for instance $x_t$ by the network $f_{\theta}(\cdot)$,  $\pi_{f_{\theta}}(x_t)= \arg \min \limits_{p \in [K]} f_{\theta}(x_t)_p$ is the treatment prescribed by the network instance $x_t$, corresponding to the lowest predicted outcome. The P-ReLU network is learned via gradient descent by minimizing equation \eqref{eq:obj_again}. To do so, we implemented the training procedure in a custom function using PyTorch. Specifically, for a given $x_t$, we calculate $f_{\theta}(x_t)$, and then out of the $K$ treatments, we select as $\pi_{f_{\theta}}(x_t)$ the index of the output that has the minimum estimated response. Then, by plugging $\pi_{f_{\theta}}(x_t)$ into the objective, we can calculate the prescriptive part of the objective function and then apply gradient descent to the total objective. 

\section{Comparison with OPTs}\label{app:compare_opt}

Conceptually, OPT imposes a tree structure a priori and estimates the counterfactuals as constant or linear functions repeatedly in every leaf during training. To estimate the treatment outcome at a leaf, only the training samples that are routed at this leaf are used to estimate the treatment outcome. As a result, when using a deeper tree, fewer samples in a leaf typically lead to higher variance in the estimates. At the end of the training, the resulting tree is a piecewise linear or piecewise constant (but not continuous) function and degenerates to a constant or linear function once a path from the root to a leaf is specified.

On the other hand, a P-ReLU is a (continuous) piecewise linear function in the domain $\mathcal{X}$ and can be converted, if needed, into a tree. P-ReLU is more expressive because i) it does not impose any particular structure on the model, ii) uses all available samples to generate the resulting piecewise linear function (and the corresponding partitions of the input space), and iii) due to the computational power of existing deep learning frameworks, large in terms of training parameters P-ReLU networks can be learned efficiently. Finally, similarly to a tree, once an activation pattern on the neurons is fixed, the network becomes a linear model on the resulting partition.

Regarding the issue of scalability, Table \ref{tab:times} summarizes the average run time taken by OPTs (OPT stands for constant predictions and axis-aligned splits; OPT-L stands for linear predictors and axis-aligned splits; OPT-L stands for constant predictions and hyperplane splits; OPT-HL stands for linear predictors and hyperplane splits but was excluded due to its prohibitive run time that was more than $14$ hours) and P-ReLU for warfarin prescription. We observe that training a P-ReLU (with 5 layers, 100 neurons per layer) is significantly faster.
\begin{table}[ht]
\centering
    \caption{Average runtime for training on the warfarin prescription problem}
\label{tab:times}
\begin{tabular}{cccc}
\toprule 
    OPT & OPT-L & OPT-H & P-ReLU \\
     \midrule
    $\approx2.6$ mins& 	$\approx 5.2$ mins&	$\approx31.6$ mins	&$\approx8$ secs\\
\bottomrule
\end{tabular} 
\end{table}
Even though OPTs are trained with coordinate descent, they do not scale when using hyperplane splits and predictors, especially for larger depth trees, since they use mixed-integer optimization to formulate the problem of finding the globally optimal prescriptive tree. Indeed, the number of binary decision variables in the mixed-integer formulation is in the order of $\mathcal{O}(2^Dn)$, while for OPTs with hyperplane splits and linear predictors the number of continuous decision variables is of $\mathcal{O}(2^Dd)$, where $D$ is the depth of the tree and $n$ is the number of samples. In addition, using linear predictors exacerbates its computational challenge because regression models need to be fit repeatedly for every leaf during each step of coordinate descent. Because of its computational bottleneck on scalability, the current OPT framework is only limited to constant and linear predictors and is unable to support more complex predictors which can improve prescriptive accuracy.

\section{Adapting Causal Trees and Forests for Multiple Treatments}\label{app:CTF}
Causal trees and forests are designed to choose between binary treatments by comparing $\mathbb{E}[Y | X = x, P = 1] - \mathbb{E} [Y | X = x, P = 0] $ to zero. To adapt CT and CF for multiple treatments with $K>2$, we incorporate the one-vs-all (1vA) and one-vs-one (1v1) schemes, similar 
to the procedures used in \citealp{kallus2017recursive}.

For 1vA, for every $p \in [K]$ we first learn an estimate of $\mathbb{E}[Y|X=x,P=p]-\mathbb{E}[Y|X=x,P\neq p]$, denoted as $\hat{\delta}^{pvA}(x)$, by applying the CT to the modified dataset $S^{pvA}_{n} = \{(X_t,\mathbbm{1}[P_t=p],y_t):t \in [n]\}$ and then we assign the treatment that does the best compared to the rest: $\hat{p}_n^{1vA}(x) \in \arg \min \limits_{p \in [K]}\hat{\delta}^{pvA}(x)$. For 1v1, there are two variants. For each $p \neq s$, we first learn an estimate of $\mathbb{E}[Y|X=x,P=p]-\mathbb{E}[Y|X=x,P=s]$, denoted as $\hat{\delta}^{pvs}(x)$, on the modified dataset $S^{pvs}_{n_p+n_s} = \{(X_t,\mathbbm{1}[P_t=p],y_t):P_t \in \{p,s\}\}$ and then we either assign the treatment that does the best compared to the worst, i.e $\hat{p}_n^{1v1-A}(x) \in \arg \min \limits_{p \in [K]}\min \limits_{s \in [K]}\hat{\delta}^{pvs}_{n_p+n_s}(x)$ or the one that gets most votes in one-to-one comparisons, $\hat{p}_n^{1v1-B}(x) \in \arg \max \limits_{p \in [K]} \sum_{p \neq s}\mathbbm{1}[\hat{\delta}^{pvs}_{n_p+n_s}(x)<0]$.

\section{Additional Details for Personalized Warfarin Dosing}

In Table \ref{tab:war_clas} we observe the mean accuracy per classifier for the multi-class classification problem with full feedback, while in Table \ref{tab:war_results_2}, we observe the mean prescription accuracy of each prescriptive algorithm, including all the variants of CT and CF, for the warfarin prescription problem.

\begin{table}[ht]
\centering
    \caption{Mean accuracy (\%) of fine-tuned classifiers for the multi-class classification problem.}
\label{tab:war_clas}
\begin{tabular}{cc}
\toprule 
    Classifier & Accuracy \\
    \midrule
    RF & $66.08 \pm 0.82$\\ 
    SVC & $67.85 \pm 0.93$ \\ 
    LogReg& $\bm{69.11} \pm 0.60$\\ 
    k-NN& $65.72 \pm 0.86$\\ 
    XGB& $67.27 \pm 0.83$\\ 
\bottomrule
\end{tabular} 
\end{table}

\begin{table}[ht]
\centering
    \caption{Mean prescription accuracy (\%) of each prescriptive algorithm for the warfarin prescription problem.}
\label{tab:war_results_2}
\begin{tabular}{cc}
\toprule 
    Algorithm & Accuracy \\
     \midrule
     R\&C RF	 & $ 65.43 \pm 1.38 $\\ 
     R\&C LR & $ 66.68\pm  0.67$\\ 
     CT 1vA & $ 47.58 \pm 2.76$\\ 
     CF 1vA & $ 64.80 \pm 0.80$\\ 
     CT 1v1-A & $ 53.70 \pm 2.31$\\
     CF 1v1-A & $ 64.61 \pm 0.69 $\\
     CT 1v1-B & $ 54.87 \pm 0.68 $\\
     CF 1v1-B & $ 54.87 \pm 0.68 $\\
     PT & $ 55.43 \pm 1.53$\\
     PF  & $ 56.03\pm 0.98 $\\
     OPT & $65.04 \pm 1.63$ \\
     P-ReLU & $ \textbf{68.27} \pm 0.76$\\
\bottomrule
\end{tabular} 

\end{table}

Before training the P-ReLU for the personalized warfarin dosing we performed standard scaling to improve the convergence of the model. The domains of the categorical features after the scaling are:
\begin{enumerate}
    \item $\texttt{Cyp2C9\_1/3}\in\{-0.31,3.24\}$.
    \item $\texttt{Cyp2C9\_1/5}\in\{-0.04,24.84\}$.
    \item $\texttt{Cyp2C9\_1/6}\in\{-0.03, 39.30\}$.
    \item $\texttt{Cyp2C9\_2/3}\in\{-0.12, 8.32\}$.
    \item $\texttt{VKORC1\_A/A}\in\{-0.63,  1.58\}$.
    \item $\texttt{Blood Clot} \in \{-0.29,  3.46\}$.
    \item $\texttt{BMI>30} \in \{0, 1\}$.
\end{enumerate}

To present the results in the main paper with the features being in a binary format, we scaled back the data to the original representation. We hereby attach the prescriptive trees with the actual cutoffs before the
reverse scaling.

\begin{figure}[ht]
\centering
\includegraphics[width=.35\linewidth]{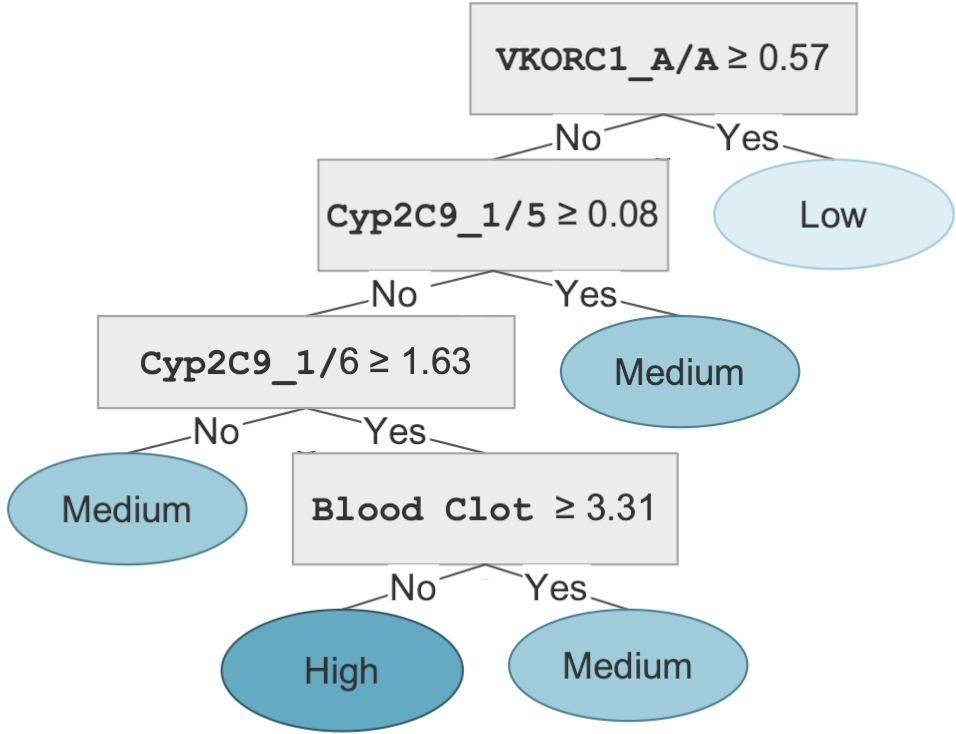}\hfill
\includegraphics[width=.6\linewidth]{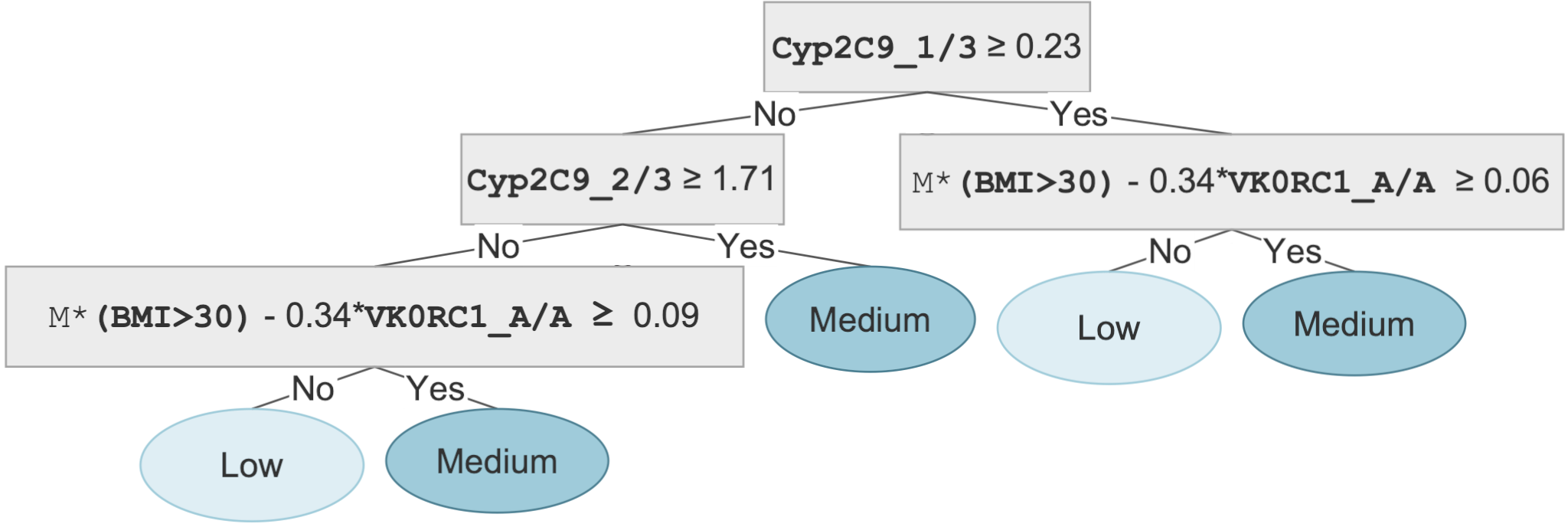}\hfill
\caption{(Left) The exact tree for the unconstrained case. (Right) The exact tree for the constrained case.}
\label{fig:exact_trees}
\end{figure}

\end{document}